\documentclass[10pt,journal,compsoc]{IEEEtran}
\usepackage[american]{babel}
\usepackage{booktabs} 
\usepackage{tikz}
\usepackage{amsmath}
\usepackage{amssymb}
\usepackage{mathtools}
\usepackage{amsthm}
\usepackage{placeins}
\usepackage{ulem}
\usepackage{graphicx}
\graphicspath{{img/}}

\usepackage{hyperref}

\begin{document}

\title{Gaussian Process-Gated Hierarchical Mixtures of Experts}
\author{Yuhao~Liu,~\IEEEmembership{Member,~IEEE,}
        Marzieh~Ajirak,~\IEEEmembership{Member,~IEEE,}
        and~Petar~M.~Djuri\'c,~\IEEEmembership{~Fellow,~IEEE}
}

\IEEEtitleabstractindextext{%
\begin{abstract}
In this paper, we propose novel Gaussian process-gated hierarchical mixtures of experts (GPHMEs). Unlike other mixtures of experts with gating models linear in the input, our model employs gating functions built with Gaussian processes (GPs). These processes are based on random features that are non-linear functions of the inputs. Furthermore, the experts in our model are also constructed with GPs. The optimization of the GPHMEs is performed by variational inference. The proposed GPHMEs have several advantages. They outperform tree-based HME benchmarks that partition the data in the input space, and they achieve good performance with reduced complexity. Another advantage is the interpretability they provide for deep GPs, and more generally, for deep Bayesian neural networks. Our GPHMEs demonstrate excellent performance for large-scale data sets, even with quite modest sizes.
\end{abstract}

}

\maketitle

\IEEEdisplaynontitleabstractindextext

\IEEEpeerreviewmaketitle

\section{Introduction}
{In this paper, we present a novel approach to constructing a hierarchical mixture of experts (HMEs) using Gaussian processes (GPs). HMEs have proven instrumental in a wide array of applications, including} regression, classification, and fusion tasks across diverse domains such as healthcare, finance, and pattern recognition \cite{jordan1994hierarchical}. {These models can be conceptualized as conditional mixture models, where the distributions of target variables are expressed as combinations of expert distributions.} The expert distributions, along with their corresponding mixing coefficients, are conditioned on the input variables. {Traditionally, model parameters are estimated by maximizing the likelihood, a practice that often leads to severe overfitting. For addressing this challenge,  a comprehensive Bayesian treatment of the model based on variational inference was proposed in \cite{bishop2012bayesian}.} 

{Our work in this paper aligns better with \cite{agrawal2021amortized}. In that paper, the authors introduced an end-to-end differentiable amortized variational inference algorithm for HMEs. The algorithm  incorporates a recurrent neural network to approximate the posterior distribution over tree node routing decisions. This not only enhances the flexibility and efficiency of the HME framework but also underscores the ongoing evolution of methodologies within this domain.}

{There is a structural similarity between hierarchical mixtures of experts and soft decision trees with fixed tree structures.} We recall that a decision tree is a hierarchical structure  composed of internal nodes and terminal leaves \cite{loh2011classification,quinlan2014c4}.  
A canonical decision tree is composed of internal nodes that represent tests on attributes. Based on the results of the test, the tree assigns samples to one of the children. The leaves on the other hand hold the labels for the classification tasks or are constants for the regression tasks. As a result, a sample traverses a single path from the root to one of the leaves.

{Decision tree nodes exhibit characteristics based on feature comparisons. Univariate nodes use a single feature, comparing it against a threshold value \cite{quinlan2014c4}.} In the case of multivariate nodes, they define a linear discriminant in the input space for comparisons \cite{murthy1994system, yildiz2005linear}, {with the ability to generalize this discriminant to be nonlinear \cite{guo1992classification}. Decision trees can implement various statistical model selection procedures at their respective nodes \cite{yildiz2001omnivariate}.}

In contrast to hard deterministic trees, where children are deterministically selected, soft probabilistic decision trees involve selecting children with specific probabilities \cite{irsoy2012soft}. {This entails traversing all the possible paths to the leaves, and the final decision is collectively contributed by all the leaves with varying probabilities.}

{The interpretability of soft decision trees extends to Gaussian soft decision trees in the context of interpretable feature-based classification \cite{yoo2021gaussian}.} In \cite{frosst2017distilling}, a deep neural network (DNN) was used to train a soft decision tree mimicking the input-output function discovered by a neural network (NN). {Soft decision trees provide interpretability for DNNs and explainability for the representations of individual hidden layers. This approach is akin to a hierarchical mixture of experts with parametric probabilistic tree-based aggregation \cite{jordan1994hierarchical}, where the experts correspond to the leaves, and the coefficients are derived from the gating nodes.}

GP extensions to HMEs have also been studied, \cite{shi2005hierarchical, ng2014hierarchical}, and in \cite{lawrence2007hierarchical}, hierarchical GP latent variable models have been investigated. We observe that the GP-based HME models rely mostly on function spaces rather than feature spaces.  


{The association between GPs and NNs has been addressed in several papers \cite{lee2017deep, wilson2020bayesian, dutordoir2021deep, pleiss2021limitations}.} Given the relationship between GPs and single-layered NNs with an infinite number of hidden units \cite{neal2012bayesian}, GPs alleviate the challenge of specifying the number of units in hidden layers by implicitly operating with infinite representations, {This is particularly beneficial in the context of feature spaces, where GP models contribute to the formulation of Bayesian Neural Networks (BNNs) capable of quantifying uncertainties.}

{Sparse GPs stand out for their ability to reduce the computational complexity of standard GPs from cubic to quadratic. Three main approaches to complexity reduction of GPs include the (1) inducing point-based approach \cite{quinonero2005unifying}, (2) finite basis expansion using Random Fourier Features (RFFs) \cite{rahimi2007random}, and (3) Hilbert space approximations of the covariance operator (HGPs) \cite{solin2020hilbert}.} 
Unlike inducing point-based methods, RFF-based GPs and HGPs do not require any matrix decomposition, and instead, they only need matrix products, which in turn boost their computational speed. These two approaches  transform nonlinear input spaces into linear kernel feature spaces, hereinafter referred to as feature spaces to distinguish them from the original input spaces.

In this paper, we work with RFF-based GPs. 
{The application of the RFF expansion to various kernels of GPs yields distinct activation functions in NN. For example, trigonometric functions are associated with the Radial Basis Function (RBF) kernel, while the ARC-COSINE kernel corresponds to Rectified Linear Unit (ReLU) functions \cite{cho2009kernel}.}  Variational learning of the posterior over the random Fourier features when the squared exponential kernel is used was proposed in \cite{gal2015improving}.

{Some recent  contributions in the field and relevant to our work include Bayesian treed GP models \cite{gramacy2008bayesian}, GP Bayesian Additive Regression Trees (GP-BART) \cite{maia2024gp}, and GP-trees \cite{achituve2021gp}. The Bayesian treed GP models allow for
nonstationary modeling that couple stationary GPs with treed partitioning. The GP-BART approach from \cite{maia2024gp} introduces a Bayesian additive regression trees methodology based on GPs. This approach is distinguished by its use of GPs within the context of additive regression trees. Namely, in standard BART, there is no covariance structure over the observations which may lead to poor performance. The GP-BART model extends BART to address this weakness by assuming GP priors for the predictions of each terminal node.
In \cite{achituve2021gp}, the authors present a GP-tree-type classifier specifically designed for few-shot incremental learning scenarios. The model addresses challenges in learning from limited labeled examples, and it works very well in scenarios where data is scarce. Other recent work in his space include \cite{achituve2021gp}.
}

The above models, partition the data only
in the input space, either in the form of HMEs or soft
decision trees. In other words, the gating models or the inner nodes are nonlinear functions of the inputs. In this paper, we propose GP-gated HMEs (GPHMEs) with structures of fixed complete binary trees where both, the gating and expert models, rely on RF expansions of GPs. This allows for making decisions by the GPs in linear feature spaces and for a considerable reduction of the complexity of the trees. We optimize the GPHMEs by using variational inference and by exploiting the reparameterization trick. In practice, the results show that the optimal height of a tree is no greater than four even for large-scale data sets with more than millions of samples. The smaller size of our tree structure is a result of the nonlinear transformation of the input space and the oblique decision mechanisms that are created. Further, the GPHME outperforms the tree-based benchmarks that partition the input data.

Our model can also be used to provide interpretability of deep GPs (DGPs). DGPs are deep belief networks based on a stack of GP mappings, where each hidden layer is composed of a multivariate GP. In practice, DGPs require less depth and width compared to DNNs. However, the DGPs encounter the problem of interpreting the hidden layers. Similar to the interpretability of DNNs distilled by soft decision trees \cite{frosst2017distilling}, this paper provides ways of getting insights into understanding how a DGP makes its decisions.


{In summary, this paper introduces a GPHME model, a novel HME that uses RF-based GPs. We also demonstrate the superiority of the GPHME over existing state-of-the-art methods.  Furthermore, the proposed method has reduced complexity and is readily applicable to large-scale problems. With our approach, we are able to quantify the uncertainty of GP probability distributions compared to trees, providing valuable insights for decision-makers. Additionally, we offer a method for interpreting DGP behaviors and offer explanations for the models located on the leaf nodes.  
Our contributions extend the current understanding of hierarchical models, offering a promising direction for future research and applications.}

\color{black}

\section{Background}
\label{sec:RF}
A GP is a stochastic process (a collection of random variables indexed by time or space), such that every finite collection of these random variables has a multivariate normal distribution. More specifically, for a finite set of inputs $\mathbf{X} \in \mathbb{R}^{N\times D_x}$, where $N$ is the number of inputs and $D_x$ is the dimension of the input, the corresponding outputs $\textbf{y} \in \mathbb{R}^{N}$ follow a GP $f$, where  $f(\mathbf{x})\sim\mathcal{GP}(\mu(\mathbf{x}), \kappa(\mathbf{x},\mathbf{x}^\prime))$, where  $\mu(\mathbf{x})$ is the mean of the process and $\kappa(\cdot,\cdot)$ is the kernel or covariance function. A popular kernel is the Radial Basis Function (RBF) defined by
\begin{equation}
\kappa(\textbf{x},\textbf{x}') = \sigma_\lambda^2\overline{\kappa}(\textbf{x},\textbf{x}') = \sigma_\lambda^2\exp\left[-\frac{1}{2}\sum_{d=1}^{D_x}\frac{(x_d-x_d')^2}{\lambda_d^2}\right],
\end{equation}
where $\sigma_\lambda^2$ is the kernel variance, {$x_d$ is the $d$-th element of $\textbf{x}$, $\lambda_d$ is the kernel lengthscale that corresponds to $x_d$}, and $\overline{\kappa}$ is the standardized kernel with norm $||\overline{\kappa}(\cdot,\cdot)||\le1$. {The parameters $\lambda _d$ are introduced to enable automated relevance determination (ARD) of the respective inputs.}

Another commonly used kernel is the ARC-COSINE kernel with a degree $n$ 
\begin{equation}
\widetilde{\kappa}(\textbf{x},\textbf{x}') =\frac{||\textbf{x}||\times||\textbf{x}'||}{\pi}J_n\left(\alpha\right),
\end{equation}
where
\begin{equation}
J_n(\alpha) = (-1)^n \left(\sin\alpha\right)^{2n+1}
\left(\frac{1}{\sin\alpha}\frac{\partial}{\partial\alpha}\right)^n\left(\frac{\pi - \alpha}{ \sin\alpha}\right),
\end{equation}
and
\begin{equation}
\alpha=\cos^{-1}\left(\frac{\textbf{x}^\top\textbf{x}'}{||\textbf{x}||||\textbf{x}'||}\right).
\end{equation}
This kernel is referred to as  an arc-cosine kernel because of its
dependence on the angle $\alpha$ and the arc-cosine function.  
In the paper, we use $n=1$ because it produces rectified linear unit (ReLU) random features {with a constant multiplier}, and 
\begin{equation}
J_1(\alpha) = \sin\alpha + (\pi - \alpha) \cos\alpha.
\end{equation}
This kernel can also implement ARD by dividing $x_d$ with $\lambda_d$.

\subsection{Random Feature Expansions for Gaussian processes}
Bochner's theorem states that if 
$\overline{\kappa}\left(\mathbf{x}_{i}, \mathbf{x}_{j}\right)=$ $\overline{\kappa}\left(\mathbf{x}_{i}-\mathbf{x}_{j}\right)$ is a continuous shift-invariant normalized covariance function, it can be rewritten as the Fourier transform of a non-negative measure $p({\boldsymbol{\omega}})$ \cite{rahimi2007random}. Let ${\boldsymbol{\omega}}$ represent a vector of spectral frequencies,  $i=\sqrt{-1}$, and $\boldsymbol{\Delta}=\mathbf{x}_{i}-\mathbf{x}_{j}$. Then we can write
\begin{equation}
\overline{\kappa}({\boldsymbol{\Delta}})=\int p(\boldsymbol{\omega}) \exp \left(i \boldsymbol{\Delta}^{\top} {\boldsymbol{\omega}}\right) d{\boldsymbol{\omega}},
\end{equation}
where $p({\boldsymbol{\omega}})$ is the Fourier transform of $\overline{\kappa}\left(\mathbf{x}_{i}, \mathbf{x}_{j}\right)$.  {
For example, if the kernel $\overline{\kappa}$ is the RBF, then the power spectral density is given by $p({\boldsymbol{\omega}})=\mathcal{N}(\textbf{0}, \Lambda^{-1})$, where $\Lambda=\text{diag} \{\lambda_1^2,...,\lambda_{D_x}^2 \}$.}

We drop the complex part of the argument of the expectation because the covariance function and the nonnegative measures are real, and we keep $\cos \left(\boldsymbol{\Delta}^{\top} {\boldsymbol{\omega}}\right)=\cos \left(\left(\textbf{x}-\textbf{x}'\right)^{\top} {\boldsymbol{\omega}}\right)$, which can also be expressed as $\cos \left(\textbf{x}^{\top} {\boldsymbol{\omega}}\right) \cos \left(\textbf{x}'^{\top} {\boldsymbol{\omega}}\right)+\sin \left(\textbf{x}^{\top} {\boldsymbol{\omega}}\right) \sin \left(\textbf{x}'^{\top} {\boldsymbol{\omega}}\right)$.
Next, we note that with the above expansion, we can estimate $\overline{\kappa}\left({\textbf{x}}, \mathbf{\textbf{x}'}\right)$   
using Monte Carlo sampling. If  $\mathbf{z}(\mathbf{\textbf{x}}, {\boldsymbol{\omega}})=$ $\left[\cos \left(\mathbf{\textbf{x}}^{\top} {\boldsymbol{\omega}}\right), \sin \left(\mathbf{\textbf{x}}^{\top} {\boldsymbol{\omega}}\right)\right]$, an unbiased estimate of $\overline{\kappa}\left(\mathbf{\textbf{x}}, \mathbf{\textbf{x}'}\right)$ can be obtained by 
\begin{equation}
\overline{\kappa}\left(\mathbf{\textbf{x}}, \mathbf{\textbf{x}'}\right) \approx \frac{1}{J} \sum_{j=1}^{J} \mathbf{z}\left(\mathbf{\textbf{x}}, {{\boldsymbol{\omega}}}_{j}\right) \mathbf{z}\left(\mathbf{\textbf{x}'}, {{\boldsymbol{\omega}}}_{j}\right)^{\top},
\end{equation}
where the ${{\boldsymbol{\omega}}}_{j}$s are samples from $p({\boldsymbol{\omega}})$, and $J$ is the number of random samples of spectral frequencies.
Using $\textbf{x}$ and the random samples $\boldsymbol{\omega}_j$, which are columns of $\boldsymbol{\Omega}$, and where  $\boldsymbol{\Omega} \in \mathbb{R}^{D_x \times J}$ (or, $\boldsymbol{\Omega}_{.j}=\boldsymbol{\omega}_j$), we define the random features {of the RBF kernel} by
\begin{equation}
\boldsymbol{\phi}_{{\text{rbf}}} (\textbf{x})\triangleq \frac{\sigma_\lambda}{\sqrt{J}}[\sin(\textbf{x}^\top\boldsymbol{\Omega}), \cos(\textbf{x}^\top\boldsymbol{\Omega})]^\top,
\end{equation}
where
\begin{align}
\sin(\textbf{x}^\top\boldsymbol{\Omega}) &=  \left[
\sin \left(\textbf{x}^\top\boldsymbol{\omega}_1 \right)\;
\sin \left(\textbf{x}^\top\boldsymbol{\omega}_2 \right)\;
\ldots
\sin \left(\textbf{x}^\top\boldsymbol{\omega}_J \right)\;
\right]
\end{align}
The definition of $\cos(\textbf{x}^\top\boldsymbol{\Omega})$ is analogous. 

For the  ARC-COSINE kernel, we can write \cite{cho2009kernel}
\begin{align}
&\widetilde{\kappa}(\textbf{x},\textbf{x}')\notag\\
&=2\int \textbf{x}^\top\boldsymbol{\omega}\;\boldsymbol{\omega}^\top\textbf{x}'\, H(\boldsymbol{\omega}^\top \textbf{x})H(\boldsymbol{\omega}^\top\textbf{x}') \mathcal{N}(\boldsymbol{\omega}|\textbf{0},\mathbf{I}) {\rm d}\boldsymbol{\omega}\notag\\
&\approx\frac{2}{J}\sum_{j=1}^J\textbf{x}^\top\boldsymbol{\omega}_j\;\boldsymbol{\omega}_j^\top\textbf{x}'\, H(\boldsymbol{\omega}_j^\top \textbf{x})H(\boldsymbol{\omega}_j^\top\textbf{x}').
\end{align}
where $H(\cdot)$ is the Heaviside function. 
Similarly to the RBF kernel, the random feature based on this covariance is defined by \cite{cutajar2017random}
\begin{equation}
\boldsymbol{\phi}_{{\text{arc}}}(\textbf{x})\triangleq\frac{\sqrt{2}\sigma_\lambda}{\sqrt{J}} \max \left(\textbf{0}, \boldsymbol{\Omega}^\top\textbf{x} \right).
\end{equation}
We note that the identity feature is given by
\begin{equation}
\label{eq:ofs}
\boldsymbol{\phi}_{{\text{I}}}(\textbf{x}) \triangleq [1\; \textbf{x}^\top]^\top.
\end{equation}

Our model reduces to an ordinary Bayesian HME when we take the feature space from \eqref{eq:ofs}. Therefore, we express the different approximation of the kernels by $k(\textbf{x},\textbf{x}')=\boldsymbol{\phi}(\textbf{x})^\top\boldsymbol{\phi}(\textbf{x}')$, and we define the GP approximation of $f$ by 
\begin{equation}
\widehat{f}(\textbf{x})=\boldsymbol{\phi}(\textbf{x})^\top\textbf{w},
\end{equation}
where the weight $\textbf{w}$ is a parameter vector with a Gaussian prior, $\mathcal{N}(\textbf{0},\mathbf{I})$.

\section{Gaussian Process-Gated Hierarchical Mixtures of Experts}
\label{sec:BSDT}

\begin{figure}[!htb]
    \centering
    \includegraphics[width=0.48\textwidth]{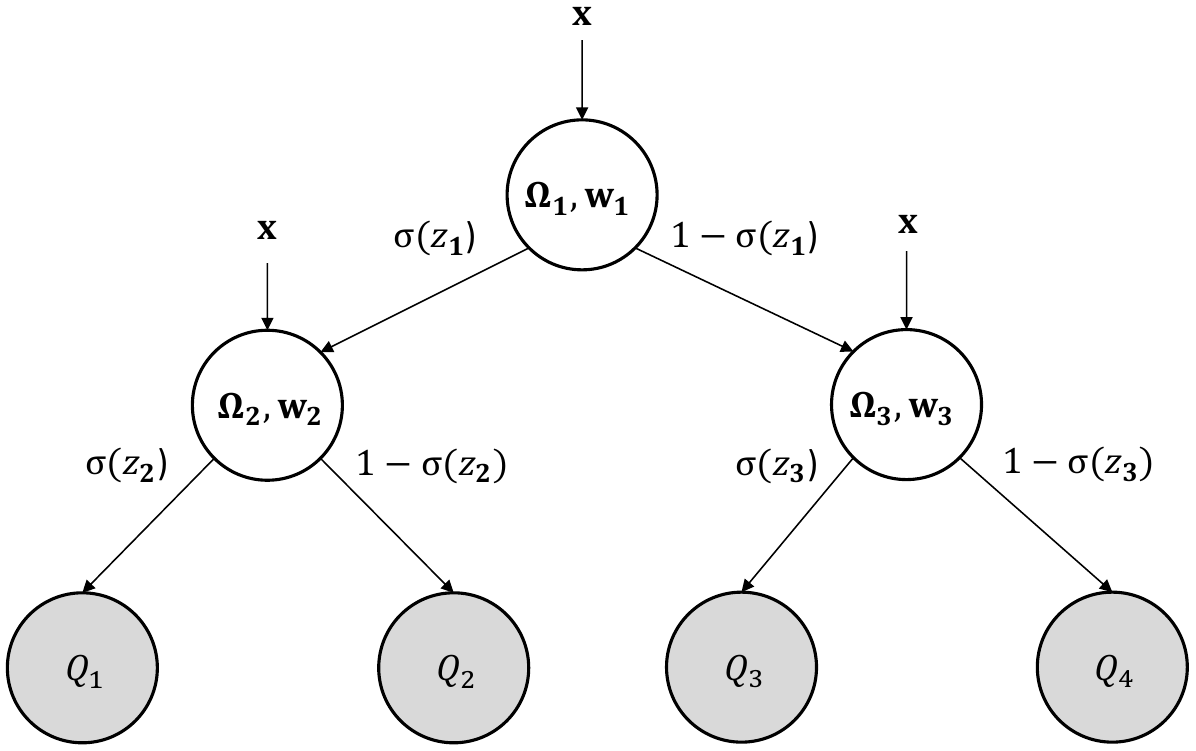}
    \caption{A GPHME with a fixed tree structure, comprising expert leaves (depicted as shaded circles) and inner nodes (depicted as circles). The edges represent RF-based decision rules associated with the inner nodes, and the $Q$s denote the conditional distributions over the target variable $y$.}
    \label{fig:bsdt}
\end{figure}

We propose GP-gated HMEs (GPHMEs) trained with mini-batch gradient descent to build mixtures of experts. An example of a GPHME is shown in Fig. \ref{fig:bsdt}. Each inner node $\nu$ of the tree has a learned set of coefficients $\textbf{w}_\nu$ and a random vector $\boldsymbol{\Omega}_\nu$, whereas each leaf node $l$ has a learned distribution $Q_l$. {At each inner node, the probability of taking the leftmost branch is $\sigma(z_\nu)$, where $\sigma(\cdot)$ is the sigmoid function.} The variable $z_\nu$ is defined by

\begin{equation}
\label{eq:node}
z_\nu(\textbf{x}) = \boldsymbol{\phi}(\textbf{x}^\top\boldsymbol{\Omega}_\nu)\textbf{w}_\nu,  
\end{equation}
where $\textbf{x}$ is the input to the model, and $\boldsymbol{\phi}$ is the random feature.
{Thus, the GP-gated model at node $\nu$ depends on $z_\nu$, which is a function of the random feature $\boldsymbol{\phi}$. Clearly, the gating probability $\sigma(z_\nu)$ is produced by first computing $z_\nu$ and then projecting it to $\left[0,1\right]$ by $\sigma(\cdot)$.}
Equation \eqref{eq:node} gives a nonlinear partition of the input space. 

The model described by Fig. \ref{fig:bsdt} is a hierarchical mixture of experts, and it learns a hierarchy of coefficients that are used to assign each sample to the GP experts with respective path probabilities. Unlike in \cite{frosst2017distilling}, where each expert is actually a ``bigot'' because it  does not use the test data to compute the experts' probability  distributions, the experts in our model learn their distributions over the possible output classes given a test  input $\textbf{x}$ according to the softmax function 
\begin{equation}
\label{eq:leaf}
Q_l^k = \frac{\exp(z_l^k)}{\sum_{k'} \exp(z_l^{k'})},
\end{equation}
where each $z_l^k=\boldsymbol{\phi}(\textbf{x}^\top\boldsymbol{\Omega}_l)\textbf{w}_l^k$ is a learned probability for class $k$ at the $l$-th leaf up to a normalizing constant, and $Q_l^k$ denotes the probability assigned to class $k$ at the $l$-th leaf. We collectively denote the hidden variables by $\boldsymbol{\Theta} = \{ \textbf{w}_\nu, \textbf{w}_l, \boldsymbol{\Omega}_\nu, \boldsymbol{\Omega}_l \}$, where the index  $\nu$ refers to the inner nodes and $l$ to the leaf nodes. The objective function, the mixture log-likelihood of $\boldsymbol{\Theta}$ for an input-output pair $(\textbf{x},y)$, is defined by
\begin{align}
\label{eq:obj_1}
{
\log p(y|\textbf{x},\boldsymbol{\Theta})} &= {\log \sum_{l}P(l|\textbf{x}, \boldsymbol{\Theta})p(y|\textbf{x},l,\boldsymbol{\Theta}) }\notag\\
&{\ge \sum_{l}P(l|\textbf{x}, \boldsymbol{\Theta})\log p(y|\textbf{x},l,\boldsymbol{\Theta})} \notag\\
&{=: \log\widetilde{p}(y|\textbf{x},\boldsymbol{\Theta}),} 
\end{align}
where $P(l|\textbf{x}, \boldsymbol{\Theta})$ is the probability of arriving at leaf node $l$ given the input $\textbf{x}$ and $\boldsymbol{\Theta}$, and $p(y|\textbf{x},l,\boldsymbol{\Theta})=Q_l^y$ is the likelihood  of $\boldsymbol{\Theta}$ of the $l$-th expert. 
{We chose to optimize $\log \widetilde{p}(y|\textbf{x},\boldsymbol{\Theta})$ over $\log p(y|\textbf{x},\boldsymbol{\Theta})$ for ELBO optimization to mitigate numerical challenges. Specifically, when $P(l|\textbf{x}, \boldsymbol{\Theta})$ and $p(y|\textbf{x},l,\boldsymbol{\Theta})$ approach zero for every leaf $l$, $\log p(y|\textbf{x},\boldsymbol{\Theta})$ tends towards negative infinity.}
For multivariate regression tasks, the leaves provide regressors simply by $Q_l^k = z_l^k$, where $k$ refers to the dimension of the multivariate ${\bf y}$. 
We reiterate that the model is by nature a hierarchical mixture of GP experts, where we project the input $\textbf{x}$ into different feature spaces to assign paths from the root to the leaves and predict the target distribution at the leaves. 

The objective function in \eqref{eq:obj_1}, however, would encourage each leave to minimize its log-likelihood, which in turn would lead each leaf to be an expert on all the classes and would result in no preference for any of the classes. In other words, the experts would become jacks of all trades.  Although this objective function can help us obtain more accurate results, we may wish that each leaf prefers a specific class to 
{make the results more explainable.}
Therefore, another option of the objective function is the normalized likelihood.
Let 
{
\begin{align}
\label{eq:repre}
\log Q^k &= \sum_{l}P(l|\textbf{x}, \boldsymbol{\Theta})\log p(k|\textbf{x},l,\boldsymbol{\Theta}) \notag\\
&\ge \sum_{l}P(l|\textbf{x}, \boldsymbol{\Theta})\log p(k|\textbf{x},l,\boldsymbol{\Theta}) \notag\\
&=: \log \widetilde{Q^k},
\end{align}
}
be proportional to the log-probability distribution of class $k$ given $\textbf{x}$. Specifically, \eqref{eq:obj_1} is equivalent to $\log Q^y$.
{Consistent with the discussion following \eqref{eq:obj_1}, we employed $\log \widetilde{Q^k}$ instead of $\log Q^k$,} and the normalized likelihood is defined by
\begin{equation}
\label{eq:obj_2}
{
\text{Normalized Likelihood} = \frac{\widetilde{Q^y}}{\sum_{k} \widetilde{Q^k}}.
}
\end{equation}
In other words, we not only maximize the likelihood of a class $y$ but also consider the likelihood of $y$ relative to other classes. Consequently, each expert would prefer one class after training, i.e., recognize one specific class with a higher probability on average with respect to other classes. We refer to the objective function in \eqref{eq:obj_1} as objective function one (OF1) and  to the objective function in \eqref{eq:obj_2} as objective function two (OF2). {We recognizes in these two functions the linear and log-linear pooling functions, respectively.} We note that multiple experts might be experts for one class and that they are trained automatically.

\section{Variational Inference}
Our goal is to find a variational distribution $q(\boldsymbol{\Theta})$ that approximates the true posterior distribution $p(\boldsymbol{\Theta}|\mathbf{X},\mathbf{Y})$. Defining the marginalized log-likelihood $L = \sum_{n=1}^N \log p(y|\textbf{x})$ and 
{
$L' = \sum_{n=1}^N \mathbb{E}_{q(\boldsymbol{\Theta})} \log p(y|\textbf{x},\boldsymbol{\Theta}) \ge  \sum_{n=1}^N \mathbb{E}_{q(\boldsymbol{\Theta})} \log \widetilde{p}(y|\textbf{x},\boldsymbol{\Theta}) =: L''$
},
we have
\begin{equation}
{
L \ge L' - \text{KL} \left[ q(\boldsymbol{\Theta})\vert\vert p(\boldsymbol{\Theta}) \right] \ge L'' - \text{KL}_{\boldsymbol{\lambda}} \left[ q(\boldsymbol{\Theta})\vert\vert p(\boldsymbol{\Theta}) \right],
}
\end{equation}
where KL stands for the Kullback-Leibler divergence, and $p(\boldsymbol{\Theta})$ is the prior distribution of the hidden variables $\boldsymbol{\Theta}$, where
\begin{align}
p(\boldsymbol{\Theta}) &=\prod_\nu \left[p(\textbf{w}_\nu) \prod_j p(\boldsymbol{\omega}_{\nu j})\right].
\end{align}
{
Through optimizing the ELBO, the hyperparameters 
in the prior $p({\boldsymbol{\omega}})$ are also optimized. 
}
Note that the KL divergence regularizes $\boldsymbol{\Theta}$ automatically, which avoids overfitting when $||\textbf{w}_\nu||$ is too large. 

We assume a Gaussian approximating distribution that factorizes across nodes. Then we have
\begin{align}
q(w_{\nu j}) &\sim \mathcal{N}(m_{\nu j}, (s_{\nu j})^2),\\
q(\Omega_{\nu ij}) &\sim \mathcal{N}(\mu_{\nu ij}, (\sigma_{\nu ij})^2),
\end{align}
where $w_{\nu j}$ represents the $j$th element of $\textbf{w}_\nu$, and $\Omega_{\nu ij}$ is the element of $\boldsymbol{\Omega}_{\nu}$ from the $i$th row and the $j$th column. The variational parameters are the mean and the variance of each of the approximating factors $m_{. j}$, $s_{. j}$, $\mu_{. ij}$, and $\sigma_{. ij}$, and we aim to optimize the lower bound with respect to these parameters. The settings of the inner nodes $\textbf{w}_\nu$ and $\boldsymbol{\Omega}_{\nu}$ are analogous to the leaves, $\textbf{w}_l^k$ and $\boldsymbol{\Omega}_l$. Because of the independent observations, we use a doubly-stochastic approximation of {$L''$}. If  we randomly select $M$ points indexed by $\mathcal{I}_M$, {$L''$} can be estimated in an unbiased way using mini-batches by
\begin{equation}
{L''} = \frac{N}{M}\sum_{m\in \mathcal{I}_M}\mathbb{E}_{q(\boldsymbol{\Theta})}L(y_m|\textbf{x}_m,\boldsymbol{\Theta}).
\end{equation}
To deal with the expectation term, we resort to Monte Carlo sampling, which yields
\begin{equation}
{L''} = \frac{N}{M}\sum_{m\in \mathcal{I}_M}\frac{1}{N_{MC}}\sum_{r=1}^{N_{MC}}L(y_m|\textbf{x}_m,\boldsymbol{\Theta}_r),
\end{equation}
where $\boldsymbol{\Theta}_r$ is sampled from $q(\boldsymbol{\Theta})$, and $N_{MC}$ is the number of drawn samples. In order to utilize the backward propagation, we apply the reparameterization trick so that the weights are reparameterized as follows:
\begin{align}
w_{\nu jr} &= m_{\nu jr} + s_{\nu jr}\epsilon_{\nu jr},\\
\Omega_{\nu ijr} &= \mu_{\nu ijr} + \sigma_{\nu ijr}\epsilon_{\nu ijr},
\end{align}
where the $\epsilon_{\nu jr}$s and $\epsilon_{\nu ijr}$s are independent samples from the standard normal distribution. 
One can also assign prior distributions to $\sigma_\lambda$ and $\lambda_d$ and then use  variational inference. 

{In essence, the leaves with GP learners compete with each other. An input $\textbf{x}$ tends to follow the path where the leaf has a more effective GP learner. The gradient descent approach compounds this tendency, leading the learning process to a state where only one leaf outperforms others significantly, and the inputs are inclined to fall into that leaf with a probability close to 1 -- a scenario similar to ``winner takes all.'' To address this issue, we adopted the solution proposed in [12], which penalizes paths to encourage each internal node to make equal use of both left and right sub-trees. We define the average probability that $\mathbf{x}$ at node $\nu$ goes to the left
child by}
\begin{equation}
\alpha_\nu = \frac{\sum_{\textbf{x}}P_\nu(\textbf{x})p_\nu(\textbf{x})}{\sum_{\textbf{x}}P_\nu({\textbf{x}})},
\end{equation}
where $P_\nu(\textbf{x})=P(\nu|\textbf{x}, \boldsymbol{\Theta})$ is the probability that $\textbf{x}$ arrives at node $\nu$ and $p_\nu(\textbf{x})$ is the probability that $\textbf{x}$ goes to the left child of node $\nu$. Therefore, the penalty term becomes
\begin{equation}
C = \lambda N\sum_{\nu} \left[0.5 \log(\alpha_\nu) + 0.5 \log(1-\alpha_\nu)\right],
\end{equation}
where $\lambda$ is a hyper-parameter and set to $2^{-d}$ in our experiments. The penalized evidence lower bound (PELBO) is given by
\begin{equation}
PELBO = {L''} + C - \text{KL} \left[ q(\boldsymbol{\Theta})\vert\vert p(\boldsymbol{\Theta}) \right].
\end{equation}

So far, we set $\boldsymbol{\Omega}_\nu$ to be different at the different nodes, i.e., the random feature spaces are unique for each node (NIS-N), and therefore, they are non-isotropic. However, this might not be necessary and could incur huge costs in space complexity. We could expect that there exists a specific random feature space that separates data linearly, which means that all the nodes share a common distribution of $\boldsymbol{\Omega}$. We refer to this option as the isotropic space of all the nodes  (ISO-N). Another option is to restrict the nodes at each level and have them share the same distribution of $\boldsymbol{\Omega}$ and thereby mitigating the computation burden. We refer to this as the isotropy of the spaces across the nodes on the same levels (ISO-L). However, this arrangement does not seem reasonable because it is difficult to justify pooling nodes from the same level that are far away from each other. In the section on numerical experiments, the results did show that this option is worse than the other options most of the time. In practice, the ISO-N option is slightly worse than NIS-N. However, despite the slightly worse performance, the ISO-N option provides better interpretability than the NIS-N option on account of the single projected feature space.

The two main operations of inference come from Eqs. \eqref{eq:node} and \eqref{eq:leaf}. For a mini batch with size $M$, the computational complexity of \eqref{eq:node} is $\mathcal{O}(MJD_xN_{MC}) + \mathcal{O}(MJN_{MC})$, and of (14), $\mathcal{O}(MJD_xN_{MC}) + \mathcal{O}(MJD_yN_{MC})$. Combined with the height of the tree $h$, the final training complexity is $\mathcal{O}(2^hMJN_{MC}(D_x+D_y))$. The test complexity is $\mathcal{O}(2^hN_{test}JN_{MC}(D_x+D_y))$, where $N_{test}$ is the size of the test set. 

The code for obtaining the results in this paper is available at Github.\footnote{{\url{https://github.com/yuhaoliu94/GP-HME}.}}

\section{Numerical Experiments}
In this section, we first show how our GPHME {interprets} the DGPs by mimicking their behaviors. Then we discuss the different settings of $\boldsymbol{\Omega}$ and choices of the height of trees. Next, we compare our methods with the Bayesian HMEs (BHMEs) \cite{bishop2012bayesian} as well as with soft and hard trees. 
{Furthermore, we conduct a comparison between our model and SOTA tree-based GPs.}
All the results suggest that there is a need for projection in feature spaces. Finally, we  show results of implementation of our methods on large-scale data.

All the experiments ran on a single machine of NVIDIA TITAN RTX GPU having 24GB RAM, but can also be directly launched on CPUs.

\subsection{{Interpretability and explainability with GPHMEs}}
\label{sec:explain}

\begin{figure*}[!tb]
    \centering
    \includegraphics[width=\textwidth]{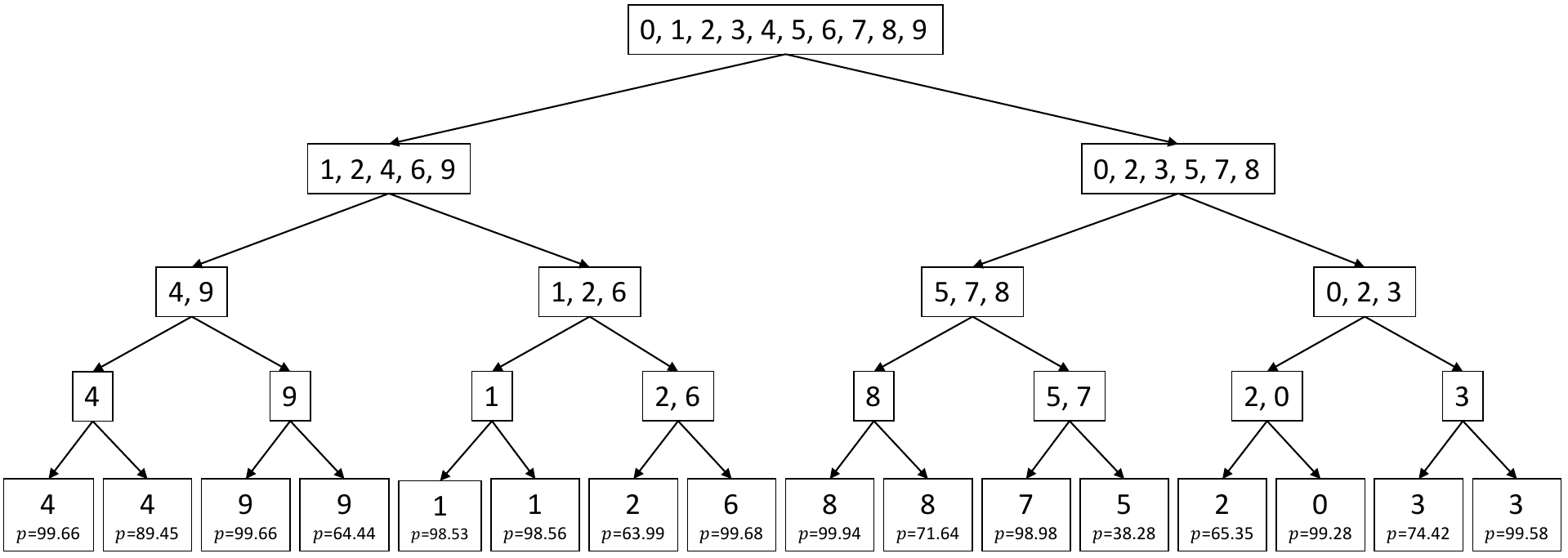}
    \caption{A visualization of a GPHME of depth four trained on the MNIST data set. The final most likely classifications are shown at each leaf with its average probability over samples. The classes annotated at each inner node are traced backward from the leaves to the root. 
    A leaf does not only predict one class but predicts all the classes. If for example, there are 100 samples and a certain leaf predicts 80 of them as digit 0 while 20 of them as any digit from 1-9, then the leaf in the figure is annotated as digit 0 with p=0.8. The classes written in the inner nodes are sourced from the leaves backward layer by layer because the predictions occur at the leaves.
    Our model is a "soft" decision tree, and the paths have probabilities, which entails that  a digit could ``go'' both left and right.}
    \label{fig:explain}
\end{figure*}

{We refer to interpretability as to using a tree-based structure to understand more complex models like DGPs and DNNs. In our approach, each decision made by a node, whether to traverse the left or right sub-tree, corresponds to a GP. Consequently, a path from the root to a leaf is established through a series of decisions, collectively forming a DGP or a DNN.}
{Within our framework, we define explainability as the ability to explain the classification results.}

Figure \ref{fig:explain} illustrates how our method makes decisions on the MNIST data. If we take, for example, the left-most four leaves, we can see that the most likely classifications are 4 and 9, and therefore, their parent node is simply learning to distinguish between these two digits. This makes sense for groups 4 and 9 because they both have closed regions in their digits. Furthermore, 2 and 5 are the most challenging for classification because none of the experts seem to reach high enough probabilities for confident decisions.   
The decisions of the inner nodes are made similarly and can readily be understood. {In summary, the explainability of our model comprises two key aspects:
\begin{itemize}
    \item Our designed normalized likelihood guides each leaf node to favor one class most, thereby explaining the function of the corresponding leaf GP.
    \item As the leaf GP predominantly favors one class, the tree structure provides an explanation of which classes demonstrate greater similarity.
\end{itemize}
}

Given the good explainability in hand, during training, our model achieves the highest accuracy of 97.79\%  with a tree of height four and with an ISO-N setting.  If we use the loss function in \eqref{eq:obj_1}, we obtain 98.49\%, still under an ISO-N setting. With more parameters, the DGP model only peaks at 98.04\% when the number of hidden layers is one and decreases with the number of layers increasing. Our GPHME model is also comparable with  a one-hidden-layer DNN, which achieves 98.4\% accuracy \cite{simard2003best}. The GPHME has this accuracy with only two-thirds of the number of parameters of the DNNs. Furthermore, the GPHME attains 98.67\% accuracy at most with a height of only two under the NIS-N setting, which is comparable to 98.6\% reported by the SVMs \cite{scholkopf1997support}. It is also better than other kernel-based methods including GPs and their variants \cite{hensman2015scalable, krauth2016autogp}.

\subsection{Discussion on hyperparameters}
\label{sec:hyper}
\begin{figure*}[!htb]
    \centering
    \includegraphics[width=\textwidth]{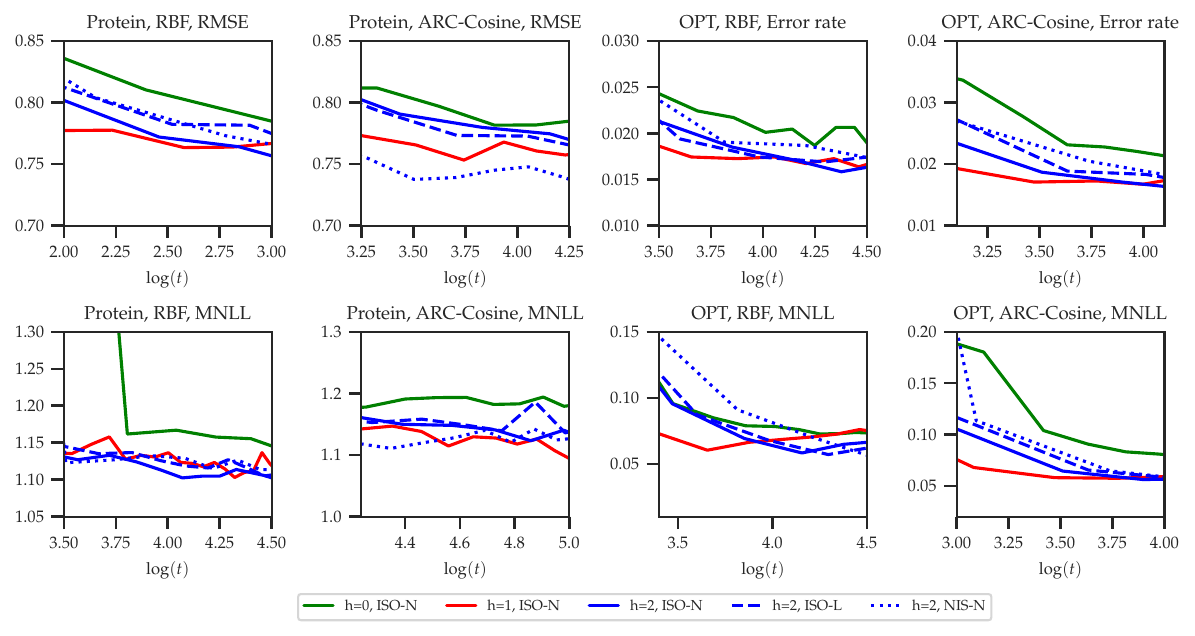}
     \caption{Evolution of RMSEs in the regression case, error rates in the classification case, and mean negative log-likelihoods (MNLLs) over time.
     The x-axes of the MNLL panels are different from those of the RMSE and ER panels because the MNLLs converge later than the RMSEs and ERs.}
    \label{fig:hyper}    
\end{figure*}

Our tree-based model, due to the information embedded in the feature spaces, does not need to grow the tree too deep.
Therefore, in this section, the heights of the trees are chosen up to two. We took one regression data set $Protein$ and one multi-output classification data set $Optical \; Digits$ (OPT) from the UCI repository.\footnote{https://archive.ics.uci.edu} The dimension of the random features was fixed to 100 for the RBF and the ARC-COSINE kernels. Figure \ref{fig:hyper} shows the performance under different settings, including different heights of trees, structures of $\boldsymbol{\Omega}$, and kernel types. Without loss of generality and making the figures readable, we only present the results of different structures of $\boldsymbol{\Omega}$ when the height of trees equals two. 

\begin{figure*}[!htb]
    \centering
    \small
    \includegraphics[width=\textwidth]{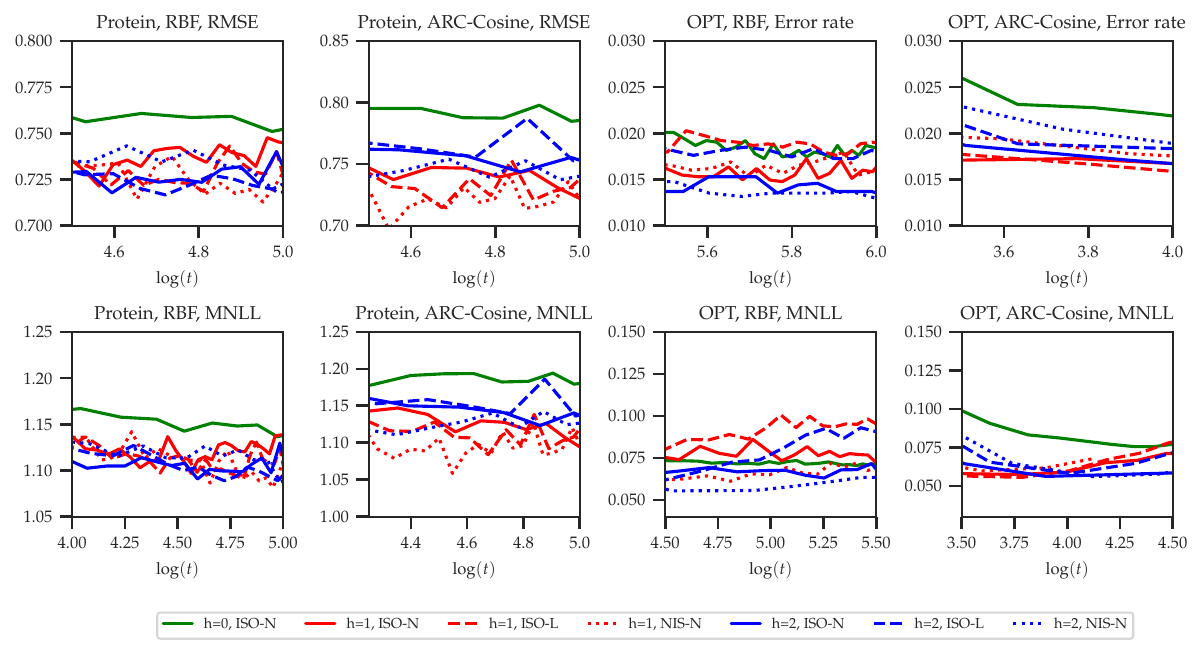}
     \caption{{Convergence} of RMSEs in the regression case, error rates in the classification case, and mean negative log-likelihoods (MNLLs) over time.}
    \label{fig:supp_hyper}
\end{figure*}

Figure \ref{fig:supp_hyper} demonstrates how the models work with different heights of trees and $\boldsymbol{\Omega}$ options. The figure also provides results under different $\boldsymbol{\Omega}$ options when the height of the tree is one.

As we expected, the performance does not always improve when the height of trees grows because of overfitting. For the setting of $\boldsymbol{\Omega}$, the NIS-N option has the most number of parameters but does not improve the results significantly compared with the ISO-L and the ISO-N options. The ISO-N option beats the NIS-N option in some cases. This is one reason why we prefer the ISO-N option in practice. The ISO-L option is not stable and might obtain the worst results among all of the $\boldsymbol{\Omega}$ settings. This is reasonable because this option forces the nodes at the same level to share the common feature spaces. As for the kernel choice, the RBF kernel has better performance in general, while the ARC-COSINE kernel may work better in binary classification tasks. The identity features are discussed in the next section as a benchmark model.

\subsection{UCI Data Sets}
{
To assess the generalization error and model complexity of various tree-based models, including ordinary decision trees (linear discriminant trees - LDT, hard tree - C4.5, and soft tree), as well as tree-based GPs (Bayesian additive regression trees - GP-BART, Bayesian treed GP - TGP, and BHMEs), we employed identical data sets as those used in \cite{irsoy2012soft}, where  results for LDT, C4.5, and soft-decision trees are reported.
} 
However, some data sets cannot be found at this time. Therefore, we only selected those that still exist, including four regression data sets (ABAlone, ADD10, BOSton, CONcrete), eight binary classification data sets (BREast, GERman, MAGic, MUSk2, PIMa, RINgnorm, SPAmbase, TWOnorm), and ten multi-class classification data sets (BALance, CMC, DERmatology, ECOli, GLAss, OPTdigits, PAGeblock, PENdigits, SEGment, YEAst) from the UCI repository. The benchmark methods use five folds of data and then average the results where one-third of the data are test data and the other two-thirds are training data. We adopted the same setting here and still let the height of the trees be at most two for our models. 
{As GP-BART \cite{maia2024gp} and TGP \cite{gramacy2008bayesian} are specifically designed for regression tasks, we evaluated their performance using regression datasets. We ensured consistency in the burn-in and total MCMC iterations for both models, following the default settings of GP-BART.}
To compare with the BHMEs fairly, we let the height of trees for BHMEs to be large enough so that the number of parameters of BHMEs is just larger than that of our models. In this case, the highest height of trees for BHMEs was usually 7 or 8 on average, although the BHMEs achieve the best results almost always with heights two or three. Besides, we provided the standard deviation across the five shuffled folds while the benchmarks of decision trees did not report this in their source. The results are selected from the best model among different heights and kernel types.

{
Tables \ref{tab:regression_removed}, \ref{tab:regression_time}, \ref{tab:regression_size} show the MSEs, training times, and tree sizes for regression tasks. Tables \ref{tab:classification_removed} and \ref{tab:multi_removed} show the accuracy of binary classifications and multi-classifications.
}
The tables also include the number of samples $N$, the dimension of inputs $D_x$, and the dimension of outputs $D_y$. 
{
The averaged training times over five different folds for classifications are also listed in Tables \ref{tab:classification_time} and \ref{tab:multi_time}.
}
The time duration for the training process was set not to be larger than 60 minutes. The sizes of the applied trees were specified by the numbers of all the inner nodes and leaves. The results of other ordinary decision trees include C4.5, which corresponds to the univariate hard tree as well as LDT, which is an oblique multivariate hard tree. 
{
The results also present additional benchmarks for tree-based GPs, including GP-BART and TGP, both utilizing deterministic linear gating rules. The resulting deterministic gating decisions led to leaves partitioning the input space into multiple regions, prompting each leaf to train a GP within its corresponding input region.
}

{
First, we compare  the results of our method with the tree-based GPs in regression, namely GP-BART and TGP. We note that our model holds advantages over GP-BART and TGP in terms of model structure and inference. The tree-based GP benchmarks used exact GPs, requiring the computation of inverse matrices of high dimensions. This approach becomes impractical when a leaf contains many samples, rendering the training time intolerable. Consequently, these methods are best suited for scarce data sets. Additionally, they applied pure MCMC approaches for inference, offering precision but at the expense of slower execution.} {Moreover, while these tree-based GP benchmarks primarily focus on explaining paths, our model goes a step further by providing explanations for the models located on the leaves. 
}

{
We conducted the unequal variance (Welch) t-test among all the models. The bolded results indicate that our method outperforms  half of the other models significantly.  From Table \ref{tab:regression_removed}, our model outperforms the benchmarks on CON data in both MSEs and training times. This is noteworthy, considering that the benchmarks are theoretically expected to be more precise due to their MCMC approaches. Referring to Table \ref{tab:regression_time}, it is evident that the benchmarks struggle to train large data sets efficiently. For ABA and ADD, the benchmarks took approximately one minute to run one iteration. The default setting requires 1,500 iterations until burn-in and 3,500 iterations to completion, requiring over two days. Note that in Table 3, tree sizes cannot be obtained from the GP-BART model's outputs.
}

{
From the results in Tables \ref{tab:classification_removed} and \ref{tab:multi_removed},
}
it is clear that our model outperforms all the other tree-based models in terms of loss. For example, for the RIN data set, the accuracy of the BHME candidates is only around $77\%$,  while our model improves it to over $98\%$ for the same size of trees. Further, our tree is smaller in size than the benchmark trees in about 80\% of the cases. We did not include the ordinary decision tree models \cite{irsoy2014budding} in the Welch t-tests because the variances of their estimates are not provided. 
The benchmark results are of trees reported in \cite{irsoy2012soft,irsoy2014budding}. However, the authors of these papers did not provide variances of the estimates.

\subsection{Large-Scale Data Sets}
The GPHMEs can be treated as weak learners and can provide interpretability for DGPs or more generally, Bayesian DNNs. We did not expect our model to perform better than DGPs but were only interested in using them to interpret DGPs. However, we were surprised that our model outperformed (at least slightly) the DGPs for most of the time.  Moreover, one of the defining characteristics of our model is the ability to scale up to large data sets. 

We evaluated our model on two large-scale problems which go beyond the scale of data sets to which GPs and especially DGPs are typically applied. We first considered MNIST8M, which artificially extends the original MNIST data set to 8+ million observations. We trained this model using the same configuration described for standard MNIST with the height of trees as two but with the NIS-N setting. We obtained $99.30\%$ accuracy and 0.0372 MNLL on the test set. These results beat the DGP counterpart provided by \cite{cutajar2017random}. We would like to point out that the number of parameters in our model is less than that of the DGPs. Note that  \cite{krauth2016autogp} recently obtained 99.11\% accuracy with the AutoGP framework with MNLL 0.033, while our model achieves the lowest MNLL 0.0328 with an accuracy of 99.24\%.

A common large-scale data in the GP field is the AIRLINE data set, which contains flight information for 5+ million US flights in 2008. Although this data set is not public, we found a substitute public data set that contains more than 6+ million records of flight information.\footnote{"https://www.kaggle.com/vikalpdongre/us-flights-data-2008"} We used this 8-dimensional data set for classification to determine whether a flight has been delayed or not. We constructed the test set using the scripts provided in \cite{wilson2016stochastic}, where 100,000 data points were held out for testing. We constructed our models using the 100 RBF-based random features and set the height of trees to one so that it matched the number of parameters in one-hidden layer DGPs \cite{cutajar2017random}. As shown in Table \ref{tab:large_scale}, the results by our model are directly comparable to those obtained by DGPs, which means that the decision tree could almost perform as well as a DGP. Further, when we grow the size of a tree to a height of four, our model achieves 72\% accuracy while the counterpart DGP with 10 hidden layers could not converge.

The training time for the MNIST data was 40 minutes, and the training time for the AIRLINE data was 19 minutes.

{Another benchmark for large-scale datasets is the GP-tree model \cite{achituve2021gp}. This model employs inducing point-based GPs as its gating model. Unfortunately, the code for this benchmark cannot be implemented successfully at present. While the code for the GP-tree model is currently unavailable for implementation, it is worth noting that our model maintains advantages over this benchmark GP-tree model.
}

{
We make two more remarks. First, the GP-tree model concentrates solely on image classification, relying on CNN for feature extraction. In contrast, our model exhibits versatility by addressing both regression and general classification tasks.
}
{
Second, the GP-tree model predefines the class labels for the leaves to ensure explainability, needing prior information on categorizing classes into distinct groups. In contrast, our model automatically directs each leaf to favor a single class.}

\section{Summary}
In this paper, we proposed a novel hierarchical mixture of experts whose inner nodes and experts are both represented by GPs. We chose to work with random features as a way of expanding the GPs. 
Our GPHMEs outperform all the benchmarks, including tree-structured HMEs and decision trees. They have reduced complexity with respect to other GP-based hierarchical mixtures of experts and offer interpretations of DGPs and deep BNNs. The HMEs have a limitation in pre-selecting the size, and the number of their parameters increases exponentially. However, it turns out that in practice, we do not need trees with large heights. Our results on various data sets clearly show excellent performance even for large-scale data sets with small trees.

Future work includes the following: 
\begin{enumerate}
\item Pruning of trees. Once the training is completed, one can proceed with  pruning the trees. For example, Fig. \ref{fig:explain} shows that we might combine the sub-trees which predict the same class. It is important to investigate principled ways of pruning. 

\item Extensions to ensembles of GPHMEs. Suppose that we have a number of GPHMEs each defined with its own set of feature spaces. Further, let these GPHMEs have prior information that their trees have structures of ISO-N, i.e., $\boldsymbol{\Omega}$ is the same across all the nodes, and they have known feature spaces $\boldsymbol{\Phi}$.  In that case, the input spaces are projected into different fixed feature spaces rather than random feature spaces. The fixed feature space is easier to interpret than random feature spaces. Therefore, the ensemble of decision trees is naturally an ensemble that exploits known fixed feature spaces. 
It is important to explore ways of identifying good ensembles, e.g., by allowing the trees to be time-varying, or by spawning new trees from the GPHMEs with good performance using distribution functions of spectral frequencies that are more informative than their respective priors.

\item Extension to boosting. This is a rather straightforward task because we only have to replicate the standard boosting routine in constructing a series of trees.   

\item Modeling experts with DGPs.  It might be possible to improve the performance of the GPHMEs by replacing the experts of the tree (the leaves), which are now GPs with DGPs. By doing this we are not losing the nice property of {explainability} of the GPHMEs but may gain in performance by using DGPs as experts.      

\item Feature Selection. A traditional tree can provide information about  the most important features of $\mathbf{X}$. Do our trees also provide information about important features? We plan to examine if information about the importance of features can be extracted from $\boldsymbol{\Omega}$ and $\textbf{w}$. 

\item Extension to inducing points approximations. With inducing points, we have an alternative approach to implementing scalable GPs. It will be interesting to examine  how trees based on such GPs compare to the ones from this paper.   

\item Feed forward option. In the paper, we only used feature spaces $\boldsymbol{\phi}(\textbf{x})$ to make decisions and predictions. However, an interesting direction is to explore the use {of} the feed-forward option for building the tree, that is, to use both $\boldsymbol{\phi}(\textbf{x})$ and $\textbf{x}$. The objective is not only to improve the accuracy of the tree in its tasks but also to understand how adding the feed-forward option affects the structure of the tree and its parameters. 

\end{enumerate}


\begin{table*}
    \caption{MSEs for regression tasks}
    \label{tab:regression_removed}
    \centering
    \small
    \begin{tabular}{r c c c c c c c c c p{0.3\textwidth}}
    \toprule
     & N & $D_x$ & GPHME & BHME & {GP-BART} & {TGP} & Soft & Hard \\
    ABA & 4177 & 8 & 0.405 $\pm$ 0.016 & 0.414 $\pm$ 0.014 & - & - & 0.439 & 0.557 \\
    ADD & 9792 & 10 & \textbf{0.043 $\pm$ 0.001} & 0.052 $\pm$ 0.003 & - & - & 0.094 & 0.267 \\
    BOS & 506 & 13 & 0.112 $\pm$ 0.023 & 0.112 $\pm$ 0.020 & {0.114 $\pm$ 0.043} & {0.101 $\pm$ 0.014} & 0.094 & 0.267\\
    CON & 1030 & 8 & \textbf{0.098 $\pm$ 0.003} & 0.147 $\pm$ 0.012 & {0.118 $\pm$ 0.019} & {0.101 $\pm$ 0.010} & 0.264 & 0.286 \\
    \bottomrule
    \end{tabular}
\end{table*}

\begin{table*}
    \caption{Training times (min) for regression tasks}
    \label{tab:regression_time}
    \centering
    \small
    \begin{tabular}{r c c c c c c c p{0.3\textwidth}}

    \toprule
     & N & $D_x$ & GPHME & BHME & GP-BART & TGP \\
    ABA & 4177 & 8 & 10.7126 & 42.9456 & $>$ 1 day & $>$ 1 day\\
    ADD & 9792 & 10 & 12.1984 & 59.9985 & $>$ 1 day & $>$ 1 day\\
    BOS & 506 & 13 & 14.8210 & 45.3653 & 2.5596 & 2.0336\\
    CON & 1030 & 8 & 12.0774 & 59.8728 & 22.2742 & 15.0907\\
    \bottomrule
    \end{tabular}
\end{table*}
\color{black}

\begin{table*}
    \caption{Tree sizes for regression tasks}
    \label{tab:regression_size}
    \centering
    \small
    \begin{tabular}{r c c c c c c c c c p{0.3\textwidth}}
    \toprule
     & N & $D_x$ & GPHME & BHME & GP-BART & TGP (max among 5 folds) & Soft & Hard \\
    ABA & 4177 & 8 & 3 & 15 & - & - & 7 & 32 \\
    ADD & 9792 & 10 & 3 & 127 & - & - & 15 & 202 \\
    BOS & 506 & 13 & 3 & 15 & - & 3 & 11 & 18\\
    CON & 1030 & 8 & 3 & 31 & - & 3 & 13 & 69 \\
    \bottomrule
    \end{tabular}
\end{table*}

\begin{table*}[!htbp]
    \caption{Accuracy for binary classification tasks}
    \label{tab:classification_removed}
    \centering
    \small
    \begin{tabular}{r c c c c c c c c c c c c c p{0.3\textwidth}}
    \toprule
     & & & \multicolumn{5}{c}{Accuracy (\%)} & \multicolumn{5}{c}{Tree Size} \\
     & N & $D_x$ & GPHME & BHME & Soft & C4.5 & LDT & GP. & BH. & Soft & C4.5 & LDT \\
    BRE & 569 & 30 & 98.42 $\pm$ 00.88 & 97.47 $\pm$ 00.91 & 95.34 & 93.80 & 95.09 & 3 & 15 & 17 & 47 & 4 \\ 
    GER & 1000 & 24 & 76.77 $\pm$ 01.42 & 76.71 $\pm$ 01.19 & 75.74 & 69.07 & 74.16 & 7 & 7 & 16 & 142 & 7 \\
    MAG & 19020 & 10 & \textbf{87.83 $\pm$ 00.22} & 79.25 $\pm$ 00.29 & 81.27 & 84.09 & 83.07 & 7 & 63 & 17 & 1072 & 40 \\
    MUS & 6598 & 167 & 100.00 $\pm$ 00.00 & 100.00 $\pm$ 00.00 & 92.25 & 94.62 & 93.59 & 3 & 3 & 22 & 202 & 15 \\
    PIM & 768 & 8 & 78.52 $\pm$ 00.96 & 78.13 $\pm$ 01.44 & 70.85 & 69.41 & 76.89 & 3 & 15 & 26 & 111 & 5 \\
    RIN & 7400 & 20 & \textbf{98.05 $\pm$ 00.16} & 76.71 $\pm$ 00.62 & 88.94 & 87.54 & 77.25 & 7 & 7 & 368 & 354 & 4 \\
    SPA & 4601 & 57 & \textbf{94.85 $\pm$ 00.41} & 92.95 $\pm$ 00.48 & 78.38 & 90.14 & 89.86 & 3 & 3 & 22 & 155 & 13 \\
    TWO & 7400 & 20 & 98.01 $\pm$ 00.24 & 98.06 $\pm$ 00.21 & 97.92 & 87.59 & 98.00 & 7 & 3 & 41 & 429 & 3 \\
    \bottomrule
    \end{tabular}
\end{table*}

\begin{table*}[!htbp]
    \caption{Accuracy for multiple classification tasks}
    \label{tab:multi_removed}
    \centering
    \small
    \begin{tabular}{r c c c c c c c c c c c c c c p{0.1\textwidth}}
    \toprule
     & & & \multicolumn{5}{c}{Accuracy(\%)} & \multicolumn{5}{c}{Tree Size} \\
     & N & $D_x$ & $D_y$ & GPHME & BHME & Soft & C4.5 & LDT & GP. & BH. & Soft & C4.5 & LDT \\
    BAL & 625 & 4 & 3 & \textbf{97.80 $\pm$ 00.72} & 90.43 $\pm$ 02.40 & 89.85 & 61.91 & 88.46 & 3 & 7 & 10 & 5 & 3 \\ 
    CMC & 1473 & 9 & 3 & 100.00 $\pm$ 00.00 & 100.00 $\pm$ 00.00 & 52.03 & 50.00 & 46.64 & 3 & 3 & 21 & 24 & 3 \\
    DER & 358 & 34 & 6 & 99.17 $\pm$ 00.53 & 98.67 $\pm$ 01.13 & 93.60 & 94.00 & 93.92 & 7 & 31 & 11 & 15 & 11 \\
    ECO & 336 & 7 & 8 & 87.86 $\pm$ 02.91 & 88.04 $\pm$ 03.37 & 76.78 & 77.47 & 81.39 & 7 & 3 & 10 & 9 & 11 \\
    GLA & 214 & 10 & 6 & 97.50 $\pm$ 01.84 & 94.72 $\pm$ 02.97 & 54.05 & 56.62 & 53.37 & 7 & 63 & 11 & 20 & 9 \\
    OPT & 5620 & 62 & 10 & \textbf{99.07 $\pm$ 00.22} & 96.95 $\pm$ 00.26 & 90.97 & 84.85 & 93.73 & 7 & 3 & 58 & 120 & 31 \\
    PAG & 5473 & 10 & 5 & 97.37 $\pm$ 00.29 & 96.59 $\pm$ 00.22 & 95.70 & 96.71 & 94.65 & 7 & 3 & 16 & 23 & 29 \\
    PEN & 10992 & 16 & 10 & \textbf{99.54 $\pm$ 00.05} & 95.53 $\pm$ 00.46 & 96.64 & 92.95 & 96.60 & 7 & 3 & 54 & 169 & 66 \\
    SEG & 2310 & 18 & 7 & \textbf{98.31 $\pm$ 00.37} & 95.35 $\pm$ 00.77 & 93.99 & 94.48 & 91.96 & 7 & 7 & 22 & 41 & 33 \\
    YEA & 1484 & 8 & 10 & 61.74 $\pm$ 00.73 & 60.00 $\pm$ 01.37 & 55.82 & 54.61 & 56.66 & 3 & 7 & 34 & 24 & 22 \\
    \bottomrule
    \end{tabular}
\end{table*}

\begin{table}[h]
    \caption{Training times (min) with 100,000 iterations for binary classifications}
    \label{tab:classification_time}
    \centering
    \small
    \begin{tabular}{r c c c c c p{0.3\textwidth}}
    \toprule
    & N & $D_x$ & GPHME Time & BHME Time \\
    BRE & 569 & 30 & 13.2187 & 52.3620 \\ 
    GER & 1000 & 24 & 28.2990 & 17.5830 \\
    MAG & 19020 & 10 & 32.0870 & 59.9671 \\
    MUS & 6598 & 167 & 14.7300 & 10.0910\\
    PIM & 768 & 8 & 17.2795 & 55.8085 \\
    RIN & 7400 & 20 & 27.1585 & 15.3693 \\
    SPA & 4601 & 57 & 16.1985 & 10.5477 \\
    TWO & 7400 & 20 & 34.2668 & 10.6366 \\
    \bottomrule
    \end{tabular}
\end{table}

\begin{table}[h]
    \caption{Training times (min) with 100,000 iterations for multiple classifications}
    \label{tab:multi_time}
    \centering
    \small
    \begin{tabular}{r c c c c c c p{0.1\textwidth}}
    \toprule
    & N & $D_x$ & $D_y$ & GPHME Time & BHME Time \\
    BAL & 625 & 4 & 3 & 16.2106 & 17.6489\\ 
    CMC & 1473 & 9 & 3 & 14.0640 & 10.9474 \\
    DER & 358 & 34 & 6 & 28.6557 & 59.9738 \\
    ECO & 336 & 7 & 8 & 32.4328 & 12.0421  \\
    GLA & 214 & 10 & 6 & 21.6130 & 59.8710  \\
    OPT & 5620 & 62 & 10 & 27.7151 & 10.0785  \\
    PAG & 5473 & 10 & 5 & 19.7589 & 11.2346 \\
    PEN & 10992 & 16 & 10 & 25.1264 & 10.2056 \\
    SEG & 2310 & 18 & 7 & 26.9447 & 15.3570 \\
    YEA & 1484 & 8 & 10 & 13.0739 & 15.3600 \\
    \bottomrule
    \end{tabular}
\end{table}

\begin{table}[!htb]
    \caption{Accuracy for large-scale data sets}
    \label{tab:large_scale}
    \centering
    \small
    \begin{tabular}{r c c c c c c p{0.25\textwidth}}
    \toprule
    & \multicolumn{2}{c}{Accuracy} & \multicolumn{2}{c}{MNLL} \\
    & GPHME & DGP & GPHME & DGP \\
    MNIST8M & 99.30\% & 99.14\% & 0.0372 & 0.0454\\
    AIRLINE & 71.79\% & 71.73\% & 0.5521 & 0.5510\\ 
    \bottomrule
    \end{tabular}
\end{table}

\FloatBarrier

\bibliographystyle{IEEEtran}
\bibliography{ref.bib, rq.bib}


\end{document}